\documentclass[conference]{IEEEtran}
\IEEEoverridecommandlockouts

\usepackage{float}  
\usepackage{algorithm}
\usepackage{algpseudocode}
\usepackage{cite}
\usepackage{amsmath,amssymb,amsfonts}
\usepackage{graphicx}
\usepackage{textcomp}
\usepackage{xcolor}
\usepackage{balance}
\usepackage{tabularx}
\usepackage{graphicx}
\usepackage{adjustbox}
\usepackage{amsmath}
\usepackage{booktabs}  
\usepackage{multirow}
\usepackage{mathrsfs}
\usepackage{enumitem}
\usepackage{stfloats}  
\usepackage{placeins} 
\usepackage{url}

\begin{document}

\title{MDD-Net: Multimodal Depression Detection through Mutual Transformer}


\author{Md Rezwanul Haque\textsuperscript{1}, Md. Milon Islam\textsuperscript{1}, S M Taslim Uddin Raju\textsuperscript{1}, Hamdi Altaheri\textsuperscript{1}, \\Lobna Nassar\textsuperscript{2}, and Fakhri Karray\textsuperscript{1,3}

\thanks{\textsuperscript{1}The authors are with the Centre for Pattern Analysis and Machine Intelligence, Department of Electrical and Computer Engineering, University of Waterloo, N2L 3G1, Ontario, Canada. (e-mail: rezwan@uwaterloo.ca, milonislam@uwaterloo.ca{*}, smturaju@uwaterloo.ca, haltaheri@uwaterloo.ca).

\textsuperscript{2}The author is with the School of Engineering and Computing, Department of Computer Science and Engineering, American University of Ras Al Khaimah, Ras Al Khaimah, United Arab Emirates. (e-mail: lobna.nassar@aurak.ac.ae).

\textsuperscript{1,3}The author is with the Centre for Pattern Analysis and Machine Intelligence, Department of Electrical and Computer Engineering, University of Waterloo, N2L 3G1, Ontario, Canada, and Department of Machine Learning, Mohamed bin Zayed University of Artificial Intelligence, Abu Dhabi, United Arab Emirates. (e-mail: karray@uwaterloo.ca, fakhri.karray@mbzuai.ac.ae).}
}
\maketitle

\begingroup
\renewcommand\thefootnote{}

\footnotetext{
\textsuperscript{*}Correspondence to: Md. Milon Islam \texttt{<milonislam@uwaterloo.ca>}.
}

\footnotetext{
\textsuperscript{©}\textit{Proceedings of the 2025 IEEE International Conference on Systems, Man, and Cybernetics (SMC), Vienna, Austria. Copyright 2025 by the author(s).}
}
\endgroup

\begin{abstract}

Depression is a major mental health condition that severely impacts the emotional and physical well-being of individuals. The simple nature of data collection from social media platforms has attracted significant interest in properly utilizing this information for mental health research. A Multimodal Depression Detection Network (MDD-Net), utilizing acoustic and visual data obtained from social media networks, is proposed in this work where mutual transformers are exploited to efficiently extract and fuse multimodal features for efficient depression detection. The MDD-Net consists of four core modules: an acoustic feature extraction module for retrieving relevant acoustic attributes, a visual feature extraction module for extracting significant high-level patterns, a mutual transformer for computing the correlations among the generated features and fusing these features from multiple modalities, and a detection layer for detecting depression using the fused feature representations. The extensive experiments are performed using the multimodal D-Vlog dataset, and the findings reveal that the developed multimodal depression detection network surpasses the state-of-the-art by up to 17.37\% for F1-Score, demonstrating the greater performance of the proposed system. The source code is accessible at \url{https://github.com/rezwanh001/Multimodal-Depression-Detection}.

\end{abstract}


\section{Introduction}

Depression is a serious psychological condition that distorts a person's mood, thoughts, and behavior. It frequently leads to sorrow, despair, and a decrease in enthusiasm for daily activities \cite{belmaker2008major}. According to the World Health Organization (WHO), approximately 322 million individuals worldwide are affected by depression, and it is expected to become the main cause of health problems by 2030 \cite{WHO2022mental}. Depression can have a profound effect on a human's life, leading to issues at work, in relationships, and even with physical health. It also imposes significant issues on society, increasing the risk of suicide and limiting social development. Currently, depression is primarily recognized through questionnaire surveys and professional assessments, however these methods can be affected by the level of participants' cooperation and clinicians' expertise \cite{yadav2023review}. Early detection and treatment are crucial, however traditional methods are often time-consuming and prone to mistakes. Therefore, developing an automated intelligent system is essential for a faster and more accurate depression detection through the use of modern technologies.

In the literature, most of the depression detection methods exploit machine learning algorithms that extract hand-crafted features for depression detection \cite{nguyen2023multimodal}. Some of the recent approaches have utilized deep learning techniques to automatically retrieve high-level attributes to efficiently detect depression \cite{tao2024depressive}. Researchers have also explored several detection methods that employ facial expressions, voice signals, and other behavioral data \cite{tao2024depmstat}. These signals are easier to capture and analyze than physiological signals such as Electrocardiography (ECG) and Electroencephalogram (EEG), which require controlled environments for data recordings. Furthermore, recent studies have shown that individuals with depression show noticeable differences in their facial expressions and speech patterns. Therefore, these expression-based techniques are also highly effective for automated depression assessment \cite{dong2021hierarchical}. 

Moreover, while most previous studies have focused on information from a single modality and considered each modality with equal importance in fusion techniques, incorporating multimodal features can significantly improve the efficiency and robustness of depression detection frameworks. This approach also makes the models more robust, allowing them to perform well even when some data types are noisy or incomplete \cite{yang2019multi}. For example, Seneviratne et al. \cite{seneviratne2022multimodal} presented a multimodal depression detection architecture combining audio and text data, addressing overfitting issues using different levels of classifiers. Yuan et al. \cite{yuan2024depression} introduced a multi-order factor fusion technique to take advantage of high-order interactions between different kinds of modalities. Shen et al. \cite{shen2022automatic} developed a depression detection system using Bi-directional Long-Short Term Memory (Bi-LSTM) and fused audio and text data with attention mechanisms. In addition, Zhou et al. \cite{zhou2022tamfn} designed the Time-Aware Attention Multimodal Fusion Network (TAMFN) to handle vlog data by overcoming issues with unimodal representations and fusion processes. The current approaches for depression detection from vlog data face challenges in extracting the most significant feature representations from multimodal data and efficiently fusing multiple features using diverse data modalities.

To address the limitations of the current depression detection models and datasets, a robust Multimodal Depression Detection Network (MDD-Net) is proposed in this work where Mutual Transformer (MT) networks are deployed to efficiently capture and fuse acoustic and visual features. The proposed approach utilizes mutual transformers to compute correlations across modalities, providing a better representation of depression-related behavioral patterns using the D-Vlog database collected from social media platforms \cite{yoon2022d}. The following are the main contributions:

\begin{itemize}
    \item  A deep learning framework with transformer architecture is developed for depression detection through multimodal data. 
  
    \item An acoustic feature extraction module is proposed, integrating a global self-attention mechanism to model content-based and positional relationships within the feature space.

    \item A video feature extraction module is introduced to extract both local and global patterns through patch embeddings and hierarchical attention mechanisms. 

    \item A mutual transformer is designed to calculate interdependencies between audio and visual embeddings, capturing multimodal information among the modalities through joint representation.

    \item The D-Vlog dataset is exploited to obtain cutting-edge performance, demonstrating the proposed model’s effectiveness in classifying between depressed and normal people based on non-verbal behavior.

\end{itemize}

The remainder of the paper is organized as follows: Section II presents a brief summary of related studies, including advances in depression detection through unimodal and multimodal approaches. Section III provides details of the proposed mutual transformer-based architecture for multimodal depression detection, highlighting its innovative fusion strategies.  Section IV discusses the D-Vlog dataset, experimental setup, results, performance comparisons, and an ablation study. Finally, Section V concludes the paper with a discussion of outcomes and recommendations for further research.

\section{Related Works}

This section reviews current approaches for depression detection. Some studies focus on short-term emotions, while others address long-term depression patterns \cite{pavlopoulos2024overview}. Researchers have utilized various methods, including Long Short-Term Memory (LSTM), Convolutional Neural Networks (CNNs), attention mechanisms, and transformers \cite{pandey2024depression, pinto2024comprehensive}. Different data modalities have been utilized, such as EEG brain signals, text, audio, video, and multimodal data \cite{zhang2024multimodal}. Video-based techniques focus on facial expressions and body language. Text-based approaches often analyze linguistic patterns, while audio-based methods leverage speech features and spectrograms. Multimodal methods, which fuse multiple data sources, have shown significant promise in enhancing recognition performance for depression detection \cite{dalal2024review}.

\subsection{Unimodal Depression Detection}

In the domain of unimodal depression detection, significant research has been conducted across various modalities to assess depressive symptoms. In \cite{deshpande2017depression}, natural language processing is utilized to detect depressive symptoms by analyzing linguistic patterns and emotional expressions in social media extracted text from platforms such as Twitter. In audio-based approaches, several techniques have been explored. Low-level descriptors and spectrogram images are commonly used as inputs to various neural network architectures to predict depression levels. For example, Kim et al. \cite{kim2023automatic} have exploited log-Mel spectrograms with a CNN architecture to predict depression. EEG brain data, often used in brain-computer interface applications, has also been used by some researchers for emotion recognition tasks \cite{erat2024emotion}. In the video-based domain, researchers have applied various visual features and deep learning techniques to detect depression. For instance, Zhou et al. \cite{zhou2018visually} have introduced DepressNet, a deep regression network designed to learn visually clear representations for depression recognition from facial expressions. The system generates a depression activation map that highlights facial regions indicative of depression severity, improving both accuracy and interpretability in depression detection.

\subsection{Multimodal Depression Detection}

Recent advancements focus on multimodal feature fusion, combining text, audio, and visual data to enhance depression detection. Fusion strategies, including early fusion, late fusion, and feature-level fusion \cite{hussain2024comprehensive}, have been employed to integrate multimodal data for enhanced depression assessment. Video content from social media platforms like YouTube has been utilized to detect depression by analyzing both audio and visual features. A technique fusing these features from video logs achieved over 75\% accuracy in recognizing depressive symptoms \cite{min2023detecting}. In another work \cite{zhou2022tamfn}, researchers introduced a method for depression detection by combining acoustic and visual data through a time-aware attention mechanism and a Temporal Convolutional Network (TCN) using the D-Vlog dataset. The system proposed a framework integrating LSTM with contextual attention and global information interaction. The proposed architecture used local information fusion to combine temporal acoustic and visual features, achieving a precision of 66.56\% on the D-Vlog dataset. A recent study by Ling et al. \cite{ling2024mdavif} presented a multimodal architecture for depression recognition from vlog data by leveraging both visual and auditory information. 

In the literature, several research studies have been conducted to efficiently detect depression from unimodal and multimodal data. However, existing frameworks face difficulties in recognizing the most relevant patterns in the raw data and developing effective feature fusion approaches to fuse them. Therefore, we propose MDD-Net for depression detection, which exploits distinct networks for audio and video to extract the most significant acoustic and visual features. Moreover, we introduce a feature fusion module based on mutual transformers to compute cross-modal correlations to generate high-level fused features.

\begin{figure*}[t]
\centerline{\includegraphics[trim={1.5cm 6.5cm 7cm 5cm},clip,scale=.60]{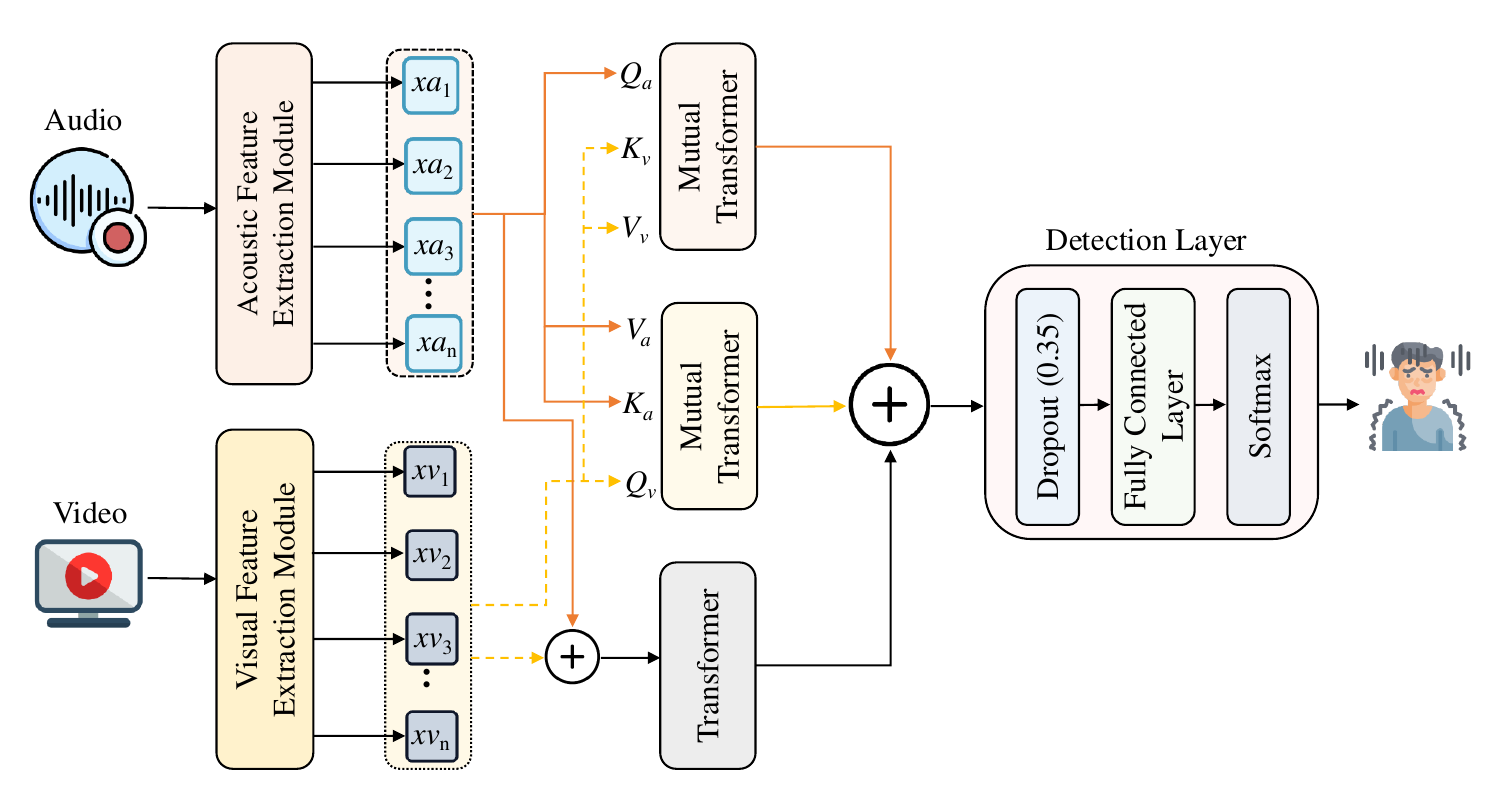}}
\caption{The overall architecture of the proposed system: a multimodal fusion technique for audio-visual feature fusion using mutual transformers for depression detection.}
\label{fig-Overall}
\end{figure*}



\begin{algorithm}[!t]
\footnotesize
\caption {Multimodal Depression Detection Network}
\begin{algorithmic}[1]
\label{alg:mdd-net}
\State \textit{\textbf{Input:}} Vlog dataset $G = \{ (X_A^n, X_V^n) \}_{n=1}^{|G|}$ 
\Statex \hspace{.77cm} where $X_A^n$ are acoustic and $X_V^n$ are visual features
\State \textit{\textbf{Output:}} Classifications $C = \{ c_n \}_{n=1}^{|G|}$
\Statex \hspace{.9cm} where $c_n \in \{ \mathrm{Depression}, \mathrm{Normal} \}$
\For{each vlog $(X_A^n, X_V^n)$ in $G$}
    \State \textbf{Extract Acoustic Features:}
    \State $X_A^o \gets \text{AFEM}(X_A^n)$ \Comment{\parbox[t]{.6\linewidth}{\scriptsize \textit{Process acoustic features using global self-attention for content and positional relationships.}}}
    \State \textbf{Extract Visual Features:}
    \State $X_V^o \gets \text{VFEM}(X_V^n)$ \Comment{\parbox[t]{.585\linewidth}{\scriptsize \textit{Process visual features with patch embedding and transformer blocks.}}}
    \vspace{.22em}
    \State \textbf{Fuse Features with Mutual Transformer:}
    \State $M C_{AV} \gets \text{MutualTransformer}_{AV}(X_A^o, X_V^o)$ \Comment{\parbox[t]{.25\linewidth}{\scriptsize \textit{Compute audio-to-video correlation.}}}
    \State $M C_{VA} \gets \text{MutualTransformer}_{VA}(X_V^o, X_A^o)$ \Comment{\parbox[t]{.25\linewidth}{\scriptsize \textit{Compute video-to-audio correlation.}}}
    \State $M C_{f_{AV}} \gets \text{TransformerFusion}(X_A^o, X_V^o)$ \Comment{\parbox[t]{.27\linewidth}{\scriptsize \textit{Fuse acoustic and visual features jointly.}}}
    \State $Z \gets \text{Combine}(M C_{AV}, M C_{VA}, M C_{f_{AV}})$ \Comment{\parbox[t]{.27\linewidth}{\scriptsize \textit{Generate multimodal representation.}}}
    \State \textbf{Detect Depression:}
    \State $p \gets \text{DetectionLayer}(Z)$ \Comment{\parbox[t] {.558\linewidth}{\scriptsize \textit{Compute probability via attention mechanism.}}}
    \If{$p > 0.5$}
        \State $c_n \gets \mathrm{Depression}$
    \Else
        \State $c_n \gets \mathrm{Normal}$
    \EndIf
\EndFor
\State \Return $C$
\end{algorithmic}
\end{algorithm}

\section{MDD-Net: Overall Architecture}
The MDD-Net architecture is designed to classify a vlog as depression or normal given a set of vlogs $G = \{g_n\}_{n=1}^{|G|}$, where each vlog $g_n$ can be represented as in \eqref{eq-g_n}.
\begin{equation}
    g_n = \left( X_{A}^{n} \in \mathbb{R}^{t \times d_a}, X_{V}^{n} \in \mathbb{R}^{t \times d_v} \right)
    \label{eq-g_n}
\end{equation}
Here, $X_{A}^{n}$ refers to the acoustic features and $X_{V}^{n}$ represents the visual features. The dimensions of the acoustic and visual features are represented by $d_a$ and $d_v$. The sequence length of the data is $t$. Both the acoustic and visual sequences are aligned to have the same length. The objective is to classify each vlog in $G$ as ``Depression'' or ``Normal'' by learning important patterns from the vlog's acoustic and visual features. The overall architecture of the proposed MDD-Net is illustrated in Fig.~\ref{fig-Overall}, and the step-by-step learning process is systematically mentioned in Algorithm 1. Each of the proposed system modules is detailed in the following sections.

\subsection{Acoustic Feature Extraction Module}

\begin{figure*}[t]
\centerline{\includegraphics[trim={0cm 12cm 1cm 7cm},clip,scale=.55]{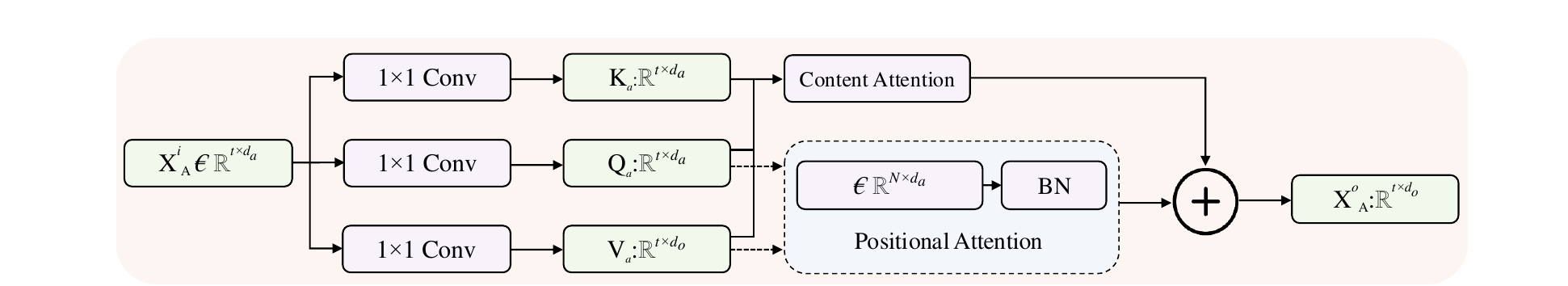}}
\caption{Acoustic feature extraction module: the module processes input audio features using $1\times1$ convolutions, applies both content and positional attention mechanisms, and performs batch normalization before fusing the features for the final representations.}
\label{fig-audio}
\end{figure*}



\begin{figure*}[b]
\centerline{\includegraphics[trim={0cm 10cm 1cm 7cm},clip,scale=.54]{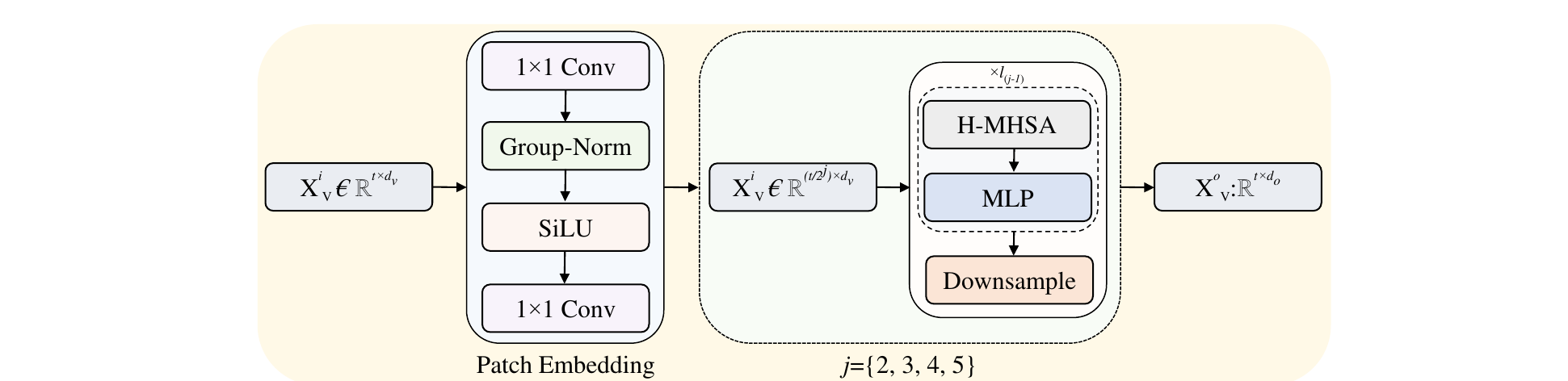}}
\caption{Visual feature extraction module: the module applies $1\times1$ convolutions, group normalization, and SiLU activation to extract patch embeddings from visual data. These embeddings are passed through a series of H-MHSA, a MLP, and a downsampling operation to generate final visual features as output.}
\label{fig-video}
\end{figure*}


In the Acoustic Feature Extraction Module (AFEM), a combination of a global self-attention network \cite{shen2020global} is used as the backbone for extracting relevant features from acoustic data. This module takes 25 low-level acoustic descriptors extracted from each audio segment, including loudness, Mel-Frequency Cepstral Coefficients (MFCCs), and spectral flux as inputs. The major components of the audio feature extraction module are represented in Fig.~\ref{fig-audio}. Let, $X_A^i \in \mathbb{R}^{t \times d_a}$ and $X_A^o \in \mathbb{R}^{t \times d_o}$ represent the input and output feature vectors, respectively. The output feature vector is captured by fusing data from the input vector utilizing both content-based and positional attention layers. The matrices of keys, queries, and values are defined as $K_a=[k_{ij}] \in \mathbb{R}^{t \times d_a}$, $Q_a = [q_{ij}] \in \mathbb{R}^{t \times d_a}$, and $V_a = [v_{ij}] \in \mathbb{R}^{t \times d_{o}}$, which are generated by applying three $1 \times 1$ convolutions on the input feature map $X_A^i$. The generated feature map $X_A^c \in \mathbb{R}^{t \times d_a}$ is then computed using content-based dot-product attention as in \eqref{eq-X_A^c}.
\begin{equation}
X_A^c = Q_a \cdot \rho(K_a^\top) \cdot V_a
\label{eq-X_A^c}
\end{equation}
where, $\rho$ represents Softmax normalization. This attention layer first aggregates the elements of $V_a$ into $d_a$ global context vectors, weighted by $\rho(K_a^\top)$, and then redistributes the context vectors to individual elements based on the weights in $Q_a$. The Softmax normalization on queries is removed to allow the output features to span the entire subspace of the context vectors $d_a$. 

To address the lack of spatial awareness in the content attention layer, a positional attention layer is introduced to handle both content and spatial relationships, as proposed in \cite{bello2019attention}. This layer analyzes the feature space in certain directions, ensuring that each element interacts with its neighbors. This approach effectively spreads information across the entire $N \times N$ neighborhood. Let, $\Delta = \left\{ -\frac{N-1}{2}, \dots, 0, \dots, \frac{N-1}{2} \right\}$ be a set of $N$ offsets, and $R = [r_\delta] \in \mathbb{R}^{N \times d_a}$ denotes the matrix of learnable relative position embeddings, where $\delta \in \Delta$. Then, our attention mechanism uses the concept of relative position embeddings ($R$) as keys, as mentioned in \eqref{eq-X_A^p}.
\begin{equation}
    X_A^{p}{}_{\left( {xy} \right)} = Q_{a \left( {xy} \right)} \cdot R^\top \cdot V_{a \left( {xy} \right)}
    \label{eq-X_A^p}
\end{equation}
where, $Q_{a \left( {xy} \right)}$ is the query, $V_{a \left( {xy} \right)} = [v_{x+\delta,y}] \in \mathbb{R}^{N \times d_{o}}$ is the matrix including the values at the $N$ neighboring elements of position $(x, y)$, and $X_A^{p}{}_{\left( {xy} \right)}$ refer to the output of the positional attention mechanism. 

The final output feature map (\(X_A^{o}\)) is a combination of these two attention mechanisms, as represented in \eqref{eq-X_A^{o}}. Here, $f_{bn}$ is the $1 \times 1$ convolution with Batch Normalization (BN) layer. 

\begin{equation}
    X_A^{o} = X_A^{c} + f_{bn}(X_A^{p})
    \label{eq-X_A^{o}}
\end{equation}

\subsection{Visual Feature Extraction Module}
For Visual Feature Extraction Module (VFEM), the network uses a patch embedding layer followed by a transformer block. The inputs of this module are 68 extracted facial landmarks for each frame in the vlog, which are fed as 136 dimensional vectors. The input feature map, $X_V^i \in \mathbb{R}^{t \times d_v}$ is processed through a $1\times1$ convolution layer, followed by group normalization, a SiLU activation function, and another $1\times1$ convolution. This acts as the patch embedding mechanism, projecting the input data into a suitable dimension for the transformer block. The video feature extraction module is represented in Fig. ~\ref{fig-video}.

Once the embedding is complete, the data is passed through transformer blocks. In the transformer block, Query ($Q_v$), Key ($K_v$), and Value ($V_v$)  are calculated using \eqref{eq-Q_v }.
\begin{equation}
    Q_v = X_V^i \beta^q, \quad K_v = X_V^i \beta^k, \quad V_v = X_V^i \beta^v
    \label{eq-Q_v }
\end{equation}
where, $ \beta^q, \beta^k, \beta^v \in \mathbb{R}^{d_v \times d_v}$ belong to the weight matrices of linear transformations during training. The Hierarchical Multi-Head Self-Attention (H-MHSA) \cite{liu2024vision} is formulated in \eqref{eq-M_V} considering the same dimensions for input and output.
\begin{equation}
    M_V = \rho \left(\frac{Q_vK_v^\top}{\sqrt{d}}\right)V_v
    \label{eq-M_V}
\end{equation}

After computing MHSA, a residual connection is incorporated to enhance optimization, as mentioned in \eqref{eq-M_V'}.
\begin{equation}
    M_V' = M_V \beta_p + X_V^i
    \label{eq-M_V'}
\end{equation}
where, $\beta_p \in \mathbb{R}^{d_v \times d_v}$  is a trainable weight matrix for feature projection. A Multi-Layer Perceptron (MLP) is applied to improve the representation, as depicted in \eqref{eq-X_V^{o}}.
\begin{equation}
    X_V^{o} = f_\text{mlp}(M_V') + M_V'
    \label{eq-X_V^{o}}
\end{equation}
where, $X_V^o$ represents the output of the transformer module. The transformer module is repeated $l_{(j-1)}$ times, where $j = \{2,3,4,5\}$.

\begin{figure}[!t]
\centerline{\includegraphics[trim={0cm 12.5cm 4.5cm 7.35cm},clip,scale=.8]{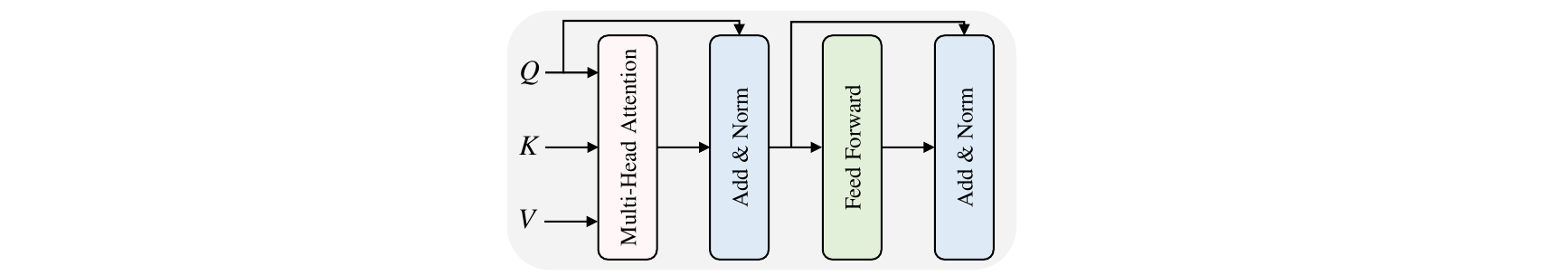}}
\caption{Transformer encoder architecture: Q, K, and V refer to input queries, keys, and values, respectively.}
\label{fig-MT}
\end{figure}

\subsection{Mutual Transformer}

The proposed approach uses the MT module to calculate multimodal mutual correlations and speaker-related depression attributes. The block diagram of the transformer is demonstrated in Fig. ~\ref{fig-MT}. From the AFEM, the acoustic features are obtained: \(X_A^o = \{x_{a1}, x_{a2}, \dots, x_{an}\}\), where \(n = 256\) refers to the sequence length and \(d_a = 71\) specifies the size for each embedding. Similarly, the visual feature extraction module provides the visual feature \(X_V^o = \{x_{v1}, x_{v2}, \dots, x_{vn}\}\), where \(n = 256\) is the length of the sequence and \(d_v = 139\) is the embedding size. The MT module bidirectionally calculates audio-to-video (\(M C_{AV}\)) and video-to-audio (\(M C_{VA}\)) correlations, followed by a joint transformer fusion. 

The Mutual Correlation (MC) is obtained using two mutual transformer layers: $MC_\text{AV} \in \mathbb{R}^{n \times d}$ and $MC_\text{VA} \in \mathbb{R}^{n \times d}$ from $X_A^o \in \mathbb{R}^{n \times d_a}$ and $X_V^o \in \mathbb{R}^{n \times d_v}$. Another transformer layer is used to fuse both $X_A^o \in \mathbb{R}^{n \times d_a}$ and $X_V^o \in \mathbb{R}^{n \times d_v}$. The mutual transformers receive query, key, and value from different modalities. In our MDD-Net, the $MC_\text{AV}$ mutual transformer receives the audio embedding $X_A^o$ as $Q_a$ and video embedding $X_V^o$ as $K_v$ and $V_v$. Similarly, $MC_\text{VA}$ receives $X_V^o$ as $Q_v$ and $X_A^o$ as $K_a$ and $V_a$. At the same time, another transformer is applied to extract the relevant features  $MC_{f_\text{AV}} \in \mathbb{R}^{(2n) \times (d_a+d_v)}$ through the fusion of $X_A^o$ and $X_V^o$. 

The corresponding formulations are presented in \eqref{eq-f_att_A_V}, \eqref{eq-MC_AV}, and \eqref{eq-MC_f_AV}.
\begin{equation}
\begin{split}
    f_{\text{attn}(X_A^o, X_V^o)} &= \rho \left( \frac{Q_a K_v^\top}{\sqrt{d_{K_v}}} \right) V_v, \\
    f_{\text{attn}(X_V^o, X_A^o)} &= \rho \left( \frac{Q_v K_a^\top}{\sqrt{d_{K_a}}} \right) V_a
\end{split}
\label{eq-f_att_A_V}
\end{equation}

\begin{equation}
\begin{split}
    MC_\text{AV} &= L\left(X_A^o + \sigma\left(\beta^\top \left(L(X_A^o + f_{\text{attn}(X_A^o, X_V^o)}) \right) \right) + b\right),\\
    MC_\text{VA} &= L\left(X_V^o + \sigma\left(\beta^\top \left(L(X_V^o + f_{\text{attn}(X_V^o, X_A^o)}) \right) \right) + b\right)
\end{split}
\label{eq-MC_AV}
\end{equation}

\begin{equation}
    MC_{f_\text{AV}} = f_\text{transformer}(f_\text{concat}(X_A^o, X_V^o))
\label{eq-MC_f_AV}
\end{equation}

where, $L$ denotes the LayerNorm. $f_{\text{attn}}$ is the self-attention function (as represented in \eqref{eq-f_att_A_V}), $f_\text{transformer}$ and $f_\text{concat}$ are the transformer and concatenation layer function of $X_A^o$ and $X_V^o$. $\beta$ and $b$ are learnable parameters.

The multimodal transformer representation \( Z \in \mathbb{R}^{(4n) \times 2(d_a + d_v)} \) is obtained using mutual transformer fusion technique. The fused output in the form of \( Z \) is shown in \eqref{eq-Z_fusion}.
\begin{equation}
    Z = f_\text{concat} \left(f_\text{AvgPool} \left(MC_\text{AV}, MC_\text{VA}, MC_{f_\text{AV}} \right) \right)
    \label{eq-Z_fusion}
\end{equation}

\subsection{Depression Detection Layer}

The output class probabilities for multimodal depression detection are derived as in \eqref{eq-P_Z_MT}. 

\begin{equation}
\begin{split}
    \alpha &= F(\beta^{\top} \cdot Z), \\
    p &= \rho \left( \beta^{\top} (\alpha^\top \cdot Z) + b \right)
\end{split}
\label{eq-P_Z_MT}
\end{equation}
Here, \( F(\cdot) \), \( \alpha \), and \( p \in \mathbb{R}^{2} \) denote the fully connected layer, attention scores, and depression detection probability, respectively. 


For depression detection, a customized loss function is utilized by combining various types of losses to address the challenges of noisy and imbalanced datasets \cite{muller2019does}. The customized loss function used in this paper combines the Binary Cross-Entropy (BCE), Focal Loss, and L2 regularization, which is written as in \eqref{eq-total_loss}.
\begin{equation}
    \mathcal{L}_{\text{total}} = \mathcal{L}_{\text{BCE}} + \mathcal{L}_{\text{Focal}} + \mathcal{L}_{\text{L2}}
    \label{eq-total_loss}
\end{equation}

The binary cross-entropy is expressed mathematically by \eqref{eq-L_BCE}, where \( y \) is the true class label and \( \epsilon_{\text{s}} \) is the smoothing factor.
\begin{equation}
    \begin{split}
    y_{\text{s}} &= {y} \cdot (1 - \epsilon_{\text{s}}) + 0.5 \cdot \epsilon_{\text{s}}, \\
    \mathcal{L}_{\text{BCE}} &= - \left( y_{\text{s}} \log(p) + (1 - y_{\text{s}}) \log(1 - p) \right)
    \label{eq-L_BCE}
    \end{split}
\end{equation}

Moreover, the Focal Loss is defined as in \eqref{eq-F_Loss}, where \( \phi \) is a scaling factor and \( \gamma \) is the focusing parameter.
\begin{equation}
    \mathcal{L}_{\text{Focal}} = \phi \cdot (1 - p)^\gamma \cdot \mathcal{L}_{\text{BCE}}
    \label{eq-F_Loss}
\end{equation}

Furthermore, the L2 regularization loss is computed as shown in \eqref{eq-L2_Loss}, where \( \theta_i \) represents the network parameters and \( \lambda \) is the strength of the L2 regularization.
\begin{equation}
    \mathcal{L}_{\text{L2}} = \lambda \sum \| \theta_i \|_2^2
    \label{eq-L2_Loss}
\end{equation}


\section{Experiments and Results Analysis}

\subsection{Dataset}
The D-Vlog dataset\cite{yoon2022d}, collected from YouTube, consists of 961 vlog videos from 816 individuals (322 men and 639 women). It includes 555 depressed and 406 normal samples. The depressed class labels are assigned based on specific keywords in the titles, with terms like  ``depression episode vlog'', ``depression daily vlog'', ``my depression story'', ``depression journey'', ``my depression diary'', ``depression video diary'', ``depression vlog''. Moreover, the non-depressed labels contain ``haul vlog'', ``daily vlog'', ``grwm (get ready with me) vlog'', ``how to vlog'', ``talking vlog'', and ``day of vlog''. The authors carried out two tasks to make sure the labels were valid: verifying that each video adheres to the vlog format with individuals speaking directly to the camera, and employing annotators to analyze videos for signs of depression using auto-generated transcripts. The database is partitioned into three sets, including training, validation, and test, with a ratio of 7:1:2. The data samples include 25 low-level acoustic descriptors from OpenSmile and 68 facial landmarks extracted by Dlib, to maintain privacy while providing relevant features.

\subsection{Experimental Setup}
All experiments are implemented using PyTorch framework in Python language. The proposed architecture is executed on a 48 GB NVIDIA RTX A6000 graphics card. The Adam optimizer is employed with the following settings: learning rate = 1e-4, weight decay = 0.1, and epsilon = 1e-8. The network is trained for 200 epochs with a batch size of 8. To prevent overfitting, an early stopping criterion is applied to stop training if no improvement in the validation error is observed for 15 epochs, saving the best model based on the validation set performance. In addition,  a 10-fold cross-validation is implemented, and the performance of the architecture is assessed using measures: Accuracy (Acc), Precision (Pr), Recall (Rc), and F1-Score (F1).

\begin{figure}[t]
\centerline{\includegraphics[trim={0 0cm 0cm 0cm},clip,scale=0.41]{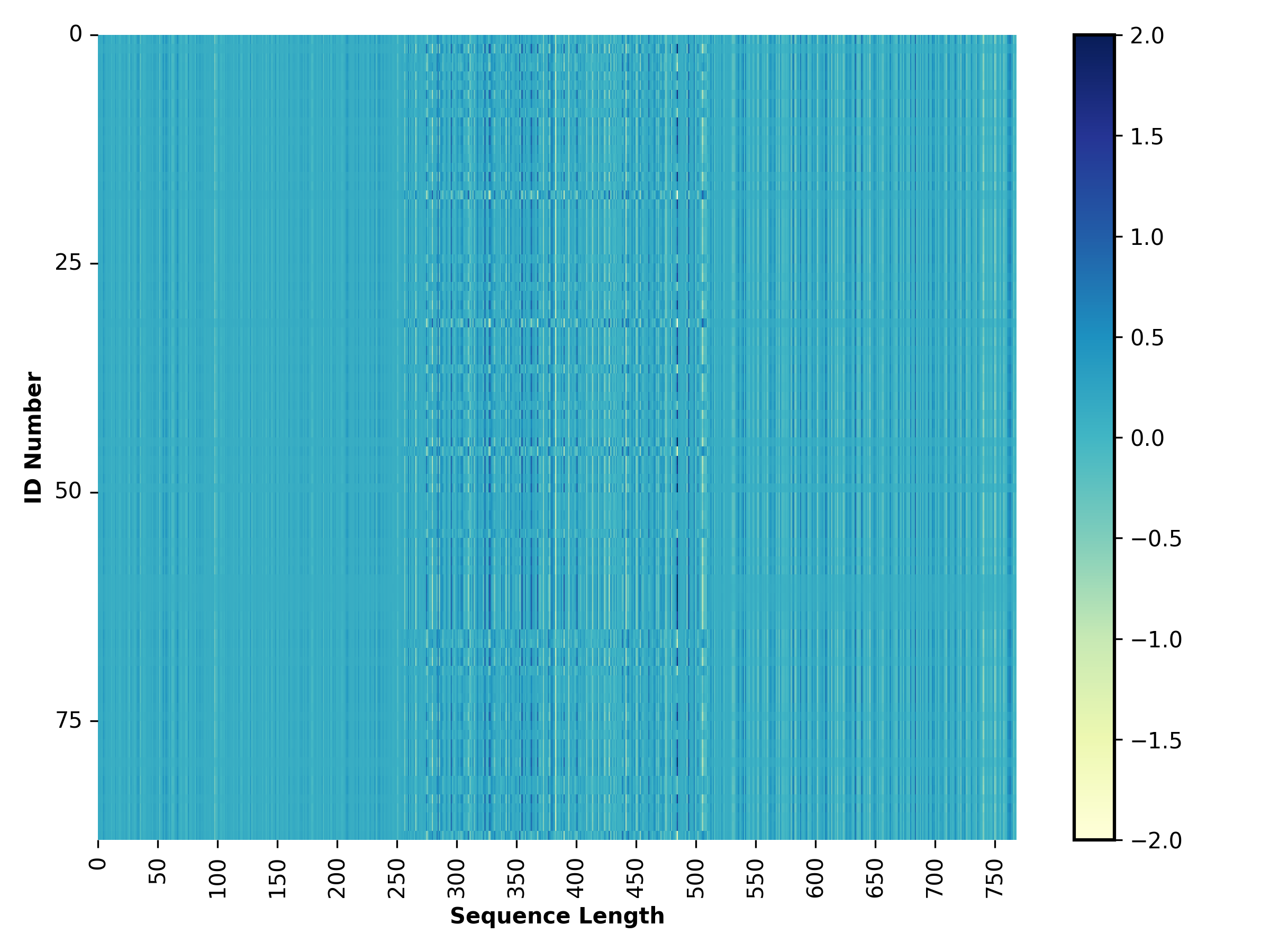}}
\centerline{(a)}
\centerline{\includegraphics[trim={0 0cm 0cm 0cm},clip,scale=0.41]{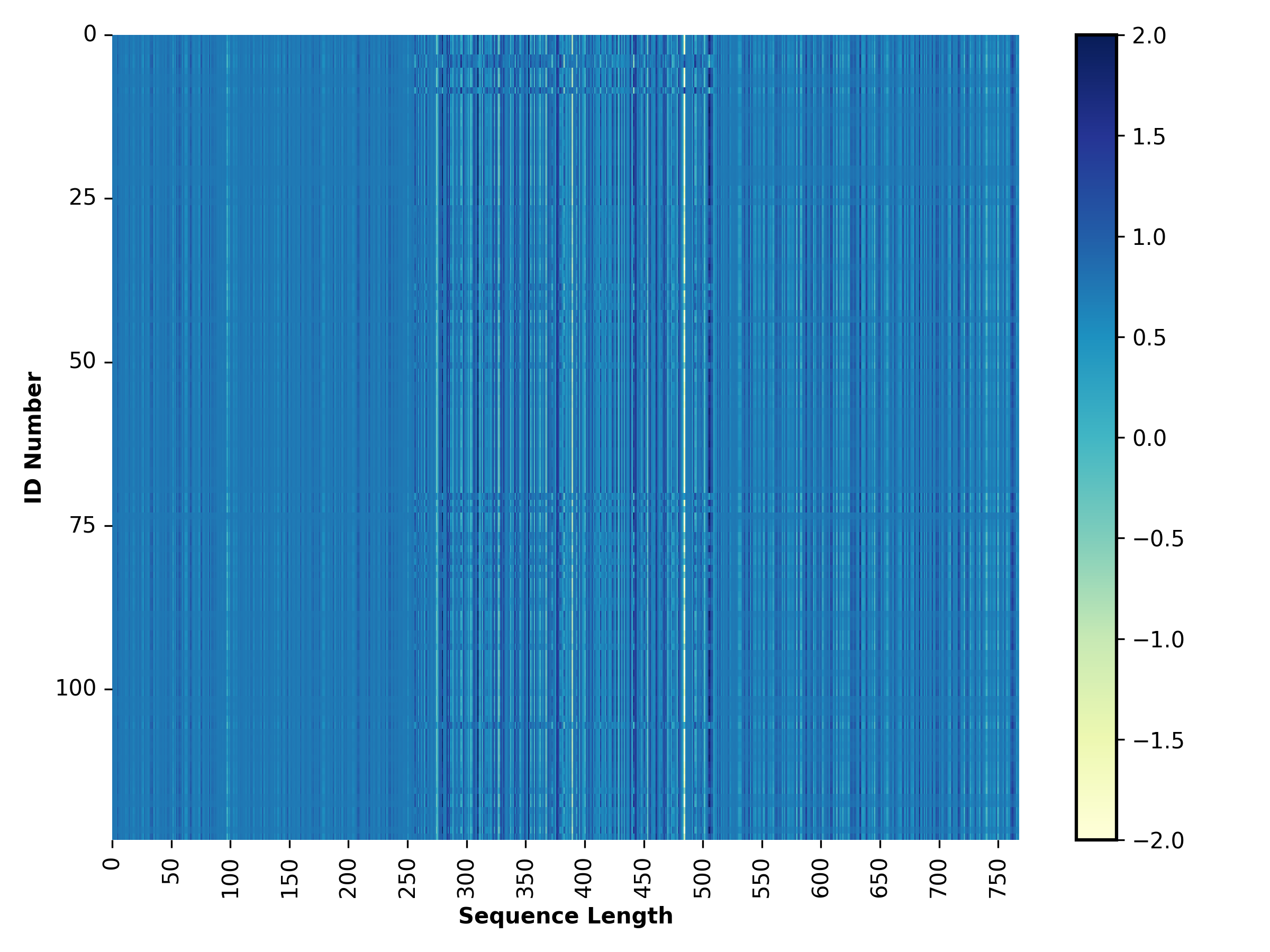}}
\centerline{(b)}
\caption{Visualization of features based on weights of the final embedding obtained by fusing audio-visual features; the darker the color, the higher the weights. (a) people with normal (b) people with depression.}
\label{fig-cam_weight}
\end{figure}


\begin{figure}[t!]
    \centering
    \includegraphics[trim={0cm 0cm 0cm 1cm}, scale=0.41]{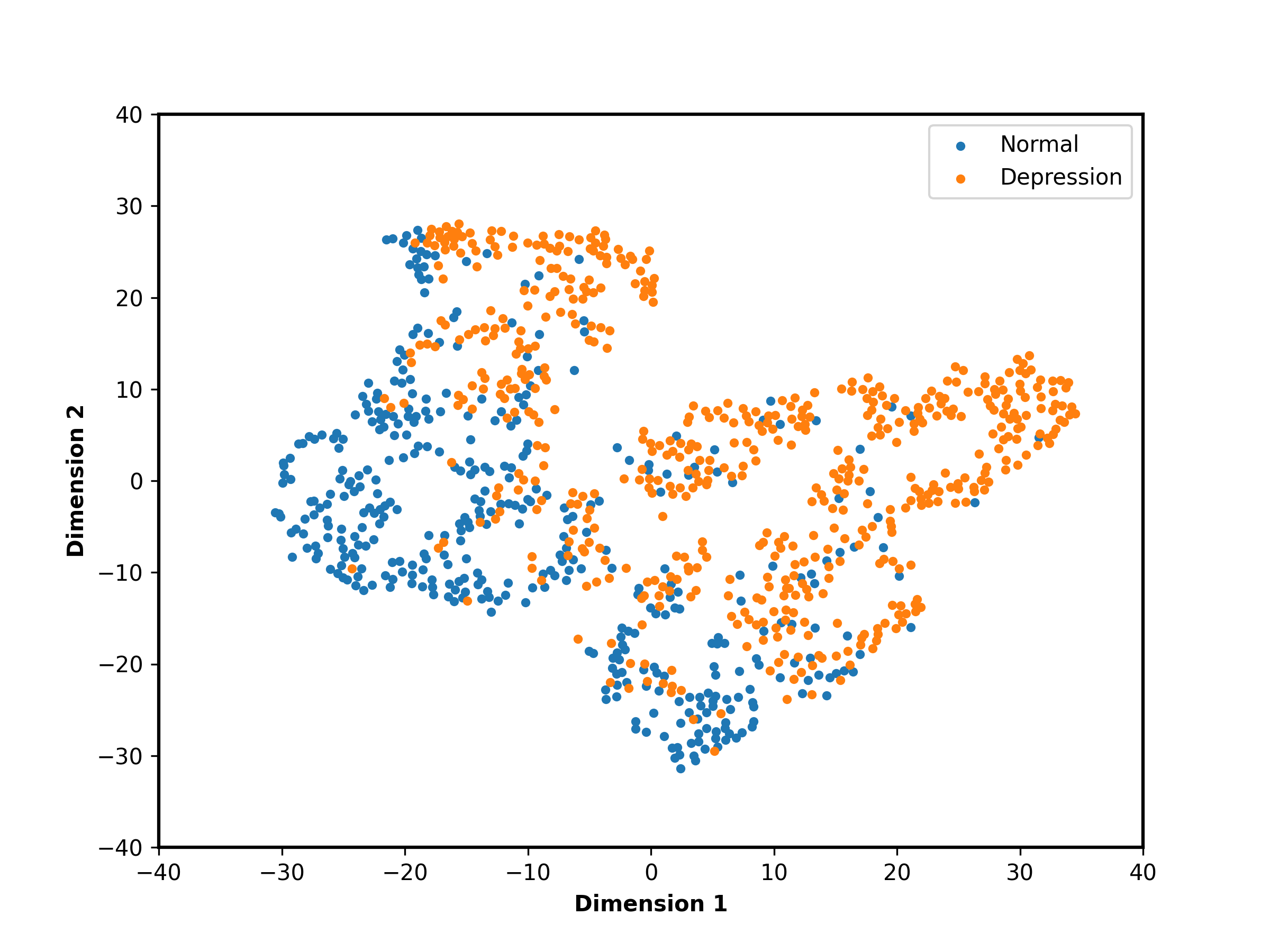}  
    \caption{Feature visualization of the proposed MDD-Net through t-SNE for multimodal depression detection.}
    \label{fig-t_sne}
\end{figure}

\subsection{Visual Results}
In this section, feature maps are presented after fusing audio and video modalities. The t-Distributed Stochastic Neighbor Embedding (t-SNE) feature visualization is utilized to provide the qualitative results \cite{dimitriadis2018t}. 

Figure~\ref{fig-cam_weight} shows the weight visualizations for both classes based on the fused acoustic-visual features. The weights are visualized on the final embeddings, which combine acoustic and visual features. Using the trained network, the weights are visualized using test samples, and the differences in sequence length between depressed and non-depressed individuals are analyzed. In Fig.~\ref{fig-cam_weight}, the darker blue colors indicate higher weights for the depressed class, while lighter blue colors refer to lower weights for the normal class.

Figure~\ref{fig-t_sne} illustrates the distribution of fused acoustic-visual features from the D-Vlog dataset using t-SNE visualization. Some normal samples overlap with the depressed category, leading to misclassifications. These misclassifications occur because the MDD-Net struggles to distinguish these categories but can still detect clusters.

\begin{table}[b]
\caption{\textsc{Performance Comparison of our Proposed MDD-Net with Different Existing Methods.}}
\begin{center}
\begin{tabular}{c c c c}
\hline
\textbf{Methods} & \textbf{Pr}& \textbf{Rc} & \textbf{F1} \\
\hline
Bi-LSTM \cite{yin2019multi} & 0.6081 & 0.6179 & 0.5970 \\

TFN \cite{zadeh2017tensor} & 0.6139 & 0.6226 & 0.6100 \\

Depression Detector \cite{yoon2022d}  & 0.6540 & 0.6557 & 0.6350 \\

TAMFN \cite{zhou2022tamfn} & 0.6602 & 0.6650 & 0.6582 \\

CAIINET \cite{zhou2023caiinet} & 0.6656 & 0.6698 & 0.6655 \\

STST \cite{tao2024depressive} & 0.7250 & 0.7767 & 0.7500 \\

MDAVIF \cite{ling2024mdavif} & \textbf{0.7425} & 0.7600 & 0.7525 \\

\textbf{MDD-Net} & 0.7392 & \textbf{0.8065} & \textbf{0.7707} \\
\hline


\end{tabular}
\label{tab-sota_res_comp}
\end{center}
\end{table}

\subsection{Results Comparison with the State-of-the-Art}
In this section, the performance of MDD-Net is compared with other methods from the literature, as shown in Table~\ref{tab-sota_res_comp}. Experimental results show that the proposed MDD-Net improves recall and F1-Score using mutual transformers by fusing features extracted from both audio and visual modules. The percentage improvement to literature for deep learning methods, such as Bi-LSTM \cite{yin2019multi} and Tensor Fusion Network (TFN) \cite{zadeh2017tensor}, ranges from 59.70\% to 61.00\% in F1-Score. More recent approaches focus on advanced feature fusion to further enhance results. For instance, the Depression Detector \cite{yoon2022d} combines audio and visual features via cross-transformers, while TAMFN \cite{zhou2022tamfn} applies time-aware attention to integrate multimodal features. CAIINET \cite{zhou2023caiinet} utilizes contextual attention and interaction to capture significant temporal information and fuse multimodal features. STST \cite{tao2024depressive} adds spatio-temporal information from various modalities for better results. Similarly, Multi-Domain Acoustical-Visual Information Fusion (MDAVIF) \cite{ling2024mdavif} fuses features from different modalities and utilizes two autoencoders to preserve information and enhance performance. The MDD-Net, proposed in this work, achieves higher performances with an enhancement of F1-Score ranging from 1.82\% to 17.37\%, with a precision of 0.7392, and recall of 0.8065, showing robust performance in detecting depression. The significant improvement of performances over single-feature modules is depicted in Table~\ref{tab-proposed_models}. The MDD- Net, with multimodal data, outperforms the acoustic feature module by 24.90\% in accuracy and 36.01\% in F1-Score. Moreover, MDD-Net shows a 26.56\% enhancement in accuracy and a 34.68\% increase in F1-Score compared to the visual feature block. The found results highlight the value of using mutual transformers to fuse features before passing them to the detection layer. The proposed fusion approach in MDD-Net architecture has proven to improve the classifier performance, making it more reliable in distinguishing between normal and depression cases across different data modalities.



\begin{table}[t!]
\caption{Performances of Proposed MDD-Net for Depression Detection from Multimodal Vlog Data.}
\begin{center}
\begin{tabular}{c c c c c c}
\hline
\textbf{Methods} & \textbf{Features} &  \textbf{Acc} & \textbf{Pr}& \textbf{Rc} & \textbf{F1} \\
\hline
AFEM & Acoustic & 0.4770 & 0.5552 & 0.4667 & 0.4106 \\

VFEM & Visual & 0.4604 & 0.5551 & 0.4577 & 0.4239 \\

MDD-Net & Both & 0.7260 & 0.7392 & 0.8065 & 0.7707 \\
\hline
\end{tabular}%
\label{tab-proposed_models}
\end{center}
\end{table}


\subsection{Ablation Study}
To test the effectiveness of each module in MDD-Net, ablation experiments are carried out, and their outcomes are analyzed in this section. To assess the impact of various fusion methods on the performance of the MDD-Net model, a series of ablation studies are performed and their findings are reported in Table \ref{tab-ablation_study}. The results show that the MT fusion method outperforms the other fusion techniques by up to 17.37\% for F1-Score, achieving the highest accuracy of 0.7260, precision of 0.7392, and recall of 0.8065. In comparison, the addition fusion (Add) method achieves an accuracy of 0.6927, with precision, recall, and F1-Score of 0.7086, 0.7909, and 0.7445, respectively. Similarly, the concatenation technique (Concat) achieves promising outcomes with an accuracy of 0.6979 and F1-Score of 0.7469, while the multiplication approach (Multiply) has a 4.06\% and 3.69\% lower performance across the accuracy and F1-Score. These findings highlight the efficiency of the MT block in capturing and fusing the interactions between acoustic and visual features, thus enhancing the network's overall performance in multimodal depression recognition. Hence, the ablation studies reveal that the design of the MT module plays a crucial role in utilizing diverse feature sets for improved classification outcomes.
\begin{table}[t!]
\caption{\textsc{Results of Ablation Studies for the Proposed MDD-Net Architecture to Detect Depression.}}
\begin{center}
\begin{tabular}{c c c c c}
\hline
\textbf{Fusion} &  \textbf{Acc} & \textbf{Pr}& \textbf{Rc} & \textbf{F1} \\
\hline
Add & 0.6927 & 0.7086 & 0.7909 & 0.7445 \\

Multiply & 0.6854 & 0.7125 & 0.7721 & 0.7338 \\
 
Concat  & 0.6979 & 0.7214 & 0.7799 & 0.7469 \\

\textbf{MDD-Net} & \textbf{0.7260} & \textbf{0.7392} & \textbf{0.8065} & \textbf{0.7707} \\
\hline

\end{tabular}
\label{tab-ablation_study}
\end{center}
\end{table}
\section{Conclusion}

Depression detection has emerged as a prominent research area due to its significant impact on people's mental and physical health condition. In this work, a multimodal depression detection system is introduced. The model utilizes mutual transformers to combine acoustic and visual features for enhancing the performance. The mutual transformer is designed to fuse the extracted features from the acoustic and visual modules for consolidating the correlation among multiple modalities. Experiments are conducted on the latest social media-based multimodal D-Vlog dataset, and the proposed network obtained comparatively better performance compared to the literature methods, with percentage improvement in F1-Score ranging from 1.82\% to 17.37\%. However, depression datasets often include limited amounts of data and imbalanced data distributions, posing difficulties for real-world applications. Given that the D-Vlog dataset is subjectively labeled by humans, mislabeling is inevitable, increasing the chance of model overfitting and limiting the network's capability to extract robust features.

In the future, more robust and reliable multimodal depression detection architectures should be developed and tested across various datasets to minimize the impact of mislabeled samples and ensure a broader applicability. Moreover, we plan to evaluate the proposed MDD-Net architecture against various fusion approaches, including late, middle, and early fusion, to assess its efficiency in enhancing cross-modal feature integration for multimodal depression detection.


\balance
\bibliographystyle{IEEEtran} 
\bibliography{references}    

\end{document}